%% file: OptionsAggregationPaper.tex
\documentclass[10pt,a4paper]{article}
\usepackage{amsmath}
\let\proof\relax
\let\endproof\relax
\usepackage{amsthm}
\usepackage{graphicx} 
\usepackage{color}

\usepackage[top=1.5cm, bottom=2cm, left=2.5cm, right=2.5cm]{geometry}

\newtheorem{obs}{Observation}

\newenvironment{proofoutline}
 {\proof[Proof outline]}
 {\endproof}

\newcommand{\argmax}{\operatornamewithlimits{argmax}}
\newcommand{\argmin}{\operatornamewithlimits{argmin}}
\newcommand{\diag}{\operatornamewithlimits{diag}}
\newcommand{\mm}[4]{\ensuremath{\left[
  \begin{array}{c|c}
    #1 & #2\\ \hline
    #3 & #4
    \end{array}\right]}}
\newcommand{\modelmatrix}[2] { \mm{1}{0}{#1}{#2} }
\newcommand{\qPonly}[0] { \Psi }
\newcommand{\qPd}[0] { \qPonly }
\newcommand{\qXa}[0] { \Xi }
\newcommand{\qProjm}[0] {\qProj^{\sminus}}
\newcommand{\qProj}[0] { \Pi }
\DeclareMathSymbol{\sminus}{\mathord}{operators}{"2D}
\newcommand{\qald}[2] { \breve{#1}^{\qPd, #2} }
\newcommand{\qal}[1] { \qald{#1}{\qXa} }

\newcommand{\qam}[0] { \qal{M} }
\newcommand{\Vb}[0] { \tilde{V} }

\begin{document}

\title{Value Iteration with Options and State Aggregation}

\author{
Kamil Ciosek \hspace{30pt} David Silver \\
Centre for Computational Statistics and Machine Learning \\
University College London
}

\maketitle

\begin{abstract}
This paper presents a way of solving Markov Decision Processes that combines state abstraction and temporal abstraction. Specifically, we combine state aggregation with the options framework and demonstrate that they work well together and indeed it is only after one combines the two that the full benefit of each is realized. We introduce a hierarchical value iteration algorithm where we first coarsely solve subgoals and then use these approximate solutions to \emph{exactly} solve the MDP. This algorithm solved several problems faster than vanilla value iteration.
\end{abstract}

\section{Introduction}
Finding solutions to discrete discounted Markov Decision Processes (MDPs) is an important problem in Reinforcement Learning. The basic problem is to obtain the optimal policy of the MDP so that the overall discounted reward obtained as we follow this policy within the MDP is maximized.\footnote{Our framework works for the discount factor $\gamma < 1$ as well as for those cases with $\gamma = 1$ where standard value termination converges (for example if there is a `sink' state).} In this work, we do not work with the optimal policy directly but compute the optimal value function instead. 

The approach we take in this paper is to modify the well-known value iteration (VI) algorithm \cite{Bellman:1957}. The basic idea of VI is to keep iterating the Bellman optimality equation. This is well-known to converge to the optimal value function. Our framework is conceptually based on a natural extension of the Bellman optimality equation where matrix models take the place of vector value functions.  

In order to solve large problems, table-lookup algorithms are not practical because of the sheer number of states, which VI must loop over. Hence the need for \emph{state abstraction}. For this work, we chose aggregation \cite{bertsekas-vol2}, which can be nicely integrated into our framework of the modified Bellman optimality equation. Algorithms based on single-step models of primitive actions are impractical, because long solution paths require many iterations of VI. Hence the need for \emph{temporal abstraction}.\footnote{Note that there is some evidence \cite{ribas2011neural} that subgoal-based hierarchical RL is similar to the processes actually taking place in the human brain.} We solve this problem via the use of options \cite{Sutton99betweenmdps,raey} --- we construct option models which can be used interchangeably with the models we have for primitive actions. 

To our knowledge, this is the first paper where an algorithm using options and value iteration efficiently solves medium-sized MDPs (our 8-puzzle domain has 181441 states). Unlike prior work \cite{options-tl}, we demonstrate a modest improvement in runtime performance as well as a significant reduction in the number of iterations. Also, we have the first \emph{convergent} VI-style algorithm where options (temporal abstraction) are combined with a framework for state abstraction, yielding far better results than the use of either idea alone. Furthermore, our algorithm is based on a principled extension of the Bellman equation. We emphasise that our algorithm converges to the \emph{optimal value function} --- although we find approximate solutions to the subgoals, these solutions are then used as inputs to solve the original MDP \emph{exactly}, regardless of the choice of subgoals. 

\section{Background and Prior Work}
\label{pw}

\subsection{State Aggregation}
\label{sec-aggr-mdp}
Consider\footnote{We refer the reader to the more elaborate introductory section in the appendix} \cite{bertsekas-vol2} an MDP with $|\mathcal{A}|$ actions; for an action $a$ the probability transition matrix is $P_a$, defined by $P_a(i,j) = \gamma \text{Pr}(i_{t+1} = j | i_{t} = i, a_t = a)$ and the vector of expected rewards for each state is $R_a$, where the element corresponding to state $i$ is defined by $R_a(i) = \text{E} \left[r_{t}  |  i_{t} = i, a_t = a \right]$. There are $m$ aggregate states. We introduce the aggregation \cite{bertsekas-vol2} matrix $\Phi$ and the disaggregation \cite{bertsekas-vol2} matrix $D$ of dimensions $n \times m$ and $m \times n$ respectively. Under the state aggregation approximation \cite{bertsekas-vol2}, solving the original MDP may be replaced by solving a much smaller aggregate MDP, by computing $\tilde{P}_a = D P_a \Phi$ and $\tilde{R}_a = D R_a$. The solution can then be computed by any known algorithm. VI is \emph{convergent} because the matrices $\tilde{P}_a$ and $\tilde{R}_a$ define a valid MDP. This gives us a value function in terms of the \emph{aggregate} states.

\subsection{Options and Matrix Models}
\label{sec-omodels}
An option \cite{Sutton95tdmodels:,Sutton99betweenmdps,raey} is a tuple $\langle \mu,\beta \rangle$, consisting of a policy $\mu$, mapping states to actions, as well as a binary termination condition $\beta$, where $\beta(i)$ tells us whether the option terminates in state $i$. We will now discuss models \cite{Sutton99betweenmdps,Sutton95tdmodels:} for options and for primitive actions. A model consists of a transition matrix $P$ and a vector of expected rewards $R$. For a primitive action $a$, we defined $P_a$ and $R_a$ in section \ref{sec-aggr-mdp}. For options they have an analogous meaning. $R(i)$ is the expected total discounted reward given the option was executed from state $i$, $R(i) = E [ \sum_{t=0}^\tau \gamma^{t} r_t | i_0 = i ]$ where $\tau$ is the (random) duration of the option and $i_0$ is the starting state. The element $P(i,i'),$ is the probability of the option terminating in state $i'$, given we started in state $i$, considering the discounting: $P(i,i') = \sum_{\tau=1}^\infty \gamma^\tau \text{Pr}(\tau, i_\tau = i' | i_0 = i)$. Denote by $i_0$ the starting state of trajectory and by $i_\tau$ the final state. It is convenient \cite{Sutton95tdmodels:} to arrange $P$ and $R$ in a block matrix of size $(n+1) \times (n+1)$, in this way: $ \modelmatrix{R}{P} $. Now model composition corresponds to matrix multiplication, i.e. if $M^{(1)}$ and $M^{(2)}$ are block matrices, $M^{(1)} M^{(2)}$ is also a block matrix corresponding to first executing the option defined in $M^{(1)}$ and then the one defined in $M^{(2)}$. In this paper, we assume that the action set $\mathcal{A} = \{ A_1, \dots, A_l \}$ is already given in this matrix format. We introduce a similar format for value functions. The value function $V$ is represented as a vector of length $n+1$ with $1$ in the first index and the values for each state in the subsequent indices. $M V$ is a new value function, corresponding to first executing the option defined in $M$ and then evaluating the states with $V$. Element $i+1$ of the vector $V$ to state $i$, as does row $i+1$ of the action model. We use MATLAB notation, i.e. $V(i+1)$ is element $i+1$ of vector $V$ and $M(i+1,:)$ is row $i+1$ of the matrix $M$.

\subsection{Other Ways of Using Hierarchies to Improve Learning}
We give a brief survey of known approaches to hierarchical learning. We stress that our approach is novel for two reasons: we \emph{compose} macro-operators at run-time and we have \emph{no fixed hierarchy}. This has not been done to date, except in the work on options and VI \cite{options-tl}, which introduced generalizations of the Bellman equation, versions of which we use. But it did not include state abstraction, slowing the resulting algorithm --- it only produced a decrease in the \emph{iteration count} required to solve the MDP, while we provide better \emph{solution time}. Other approaches include using macro-operators to gain speed in planning \cite{korf1985learning}, but for deterministic systems only. Prior work on HEXQ \cite{hengst2002discovering} is largely orthogonal to ours -- it focuses on hierarchy discovery, while we describe an algorithm \emph{given} the subgoals. The work on portable options \cite{konidaris2007building} only discusses a flat, fixed (unlike this work) options hierarchy. MAXQ \cite{dietterich1998maxq} also involves a pre-defined controller hierarchy (the MAXQ graph)\footnote{One can learn a MAXQ hierarchy \cite{wang-maxq}, but only in a way when it is first learned and then applied.}. Combining the use of temporal and state abstraction was tried before, but differently from this work. The abstraction-via-statistical-testing approach \cite{jong2005state} only works for transfer learning --- options are only constructed after the original MDP has been solved. The U-tree approach \cite{jonsson2001automated} does not guarantee convergence to $V^\star$ for all MDPs. The modified LISP approach \cite{andre2002state} uses a fixed option hierarchy and the policy obtained is only optimal given the hierarchy, i.e.\ it may not be the optimal policy of the MDP without the hierarchy constraint.

\section{Table-lookup Value Iteration}
\label{iteration}
We begin by describing the table-lookup algorithm for computing the value function of the MDP. It is similar to the one described in previous work \cite{options-tl}, but not the same --- here, termination is implemented in a different, more intuitive, way. We start with plain VI and then proceed to more complicated variants. In MATLAB notation (see section \ref{sec-omodels}), VI can be described as follows for state $i$.
\begin{equation}
\label{eq-vi}
V_{(k+1)}(i+1) \leftarrow \max_a A_a(i+1,:) V_{(k)}
\end{equation}
Here, $a$ selects an action (control). We rewrite this update  to construct a model corresponding to the optimal value function --- this is not useful on its own, but will come in handy later. The following is executed for each state $i$.
\begin{gather}
\label{eq-vi-model}
a \leftarrow \argmax_a A_a(i+1,:) M_{(k)} [1, 0, \dots, 0]^\top; \nonumber \\
M_{(k+1)}(i+1,:) \leftarrow A_a(i+1,:) M_{(k)} 
\end{gather}
We note that the multiplication $M_{(k)} [1, 0, \dots, 0]^\top$ simply extracts the total reward in the model $M_{(k)}$ (the current value function) --- hence eq.\ \ref{eq-vi-model} is equivalent to plain VI. However, it serves an an important stepping stone to introducing \emph{subgoals}, which is what we do next. Assume that we are, for the moment, not interested in maximizing the overall reward. Instead, we want to reach some other arbitrary configuration of states defined by the subgoal vector $G$ of length $n+1$. The entry $i+1$ of $G$ defines the value we associate with reaching state $i$. We will show later how picking such subgoals judiciously can improve the speed of the overall algorithm. Our new update, for subgoal $G$ is the following, which we execute for each state $i$.
\begin{gather}
a \leftarrow \argmax_a A_a(i+1,:) M_{(k)} G; \nonumber \\
M_{(k+1)}(i+1,:) \leftarrow A_a(i+1,:) M_{(k)}
\end{gather}
This iteration converges \cite{options-tl} to a model $M_\infty$, which corresponds to the policy for reaching the subgoal $G$. However, this policy executes continually, it does not stop when a state with a high subgoal value of $G(i+1)$ is reached. We will now fix that by introducing the possibility of termination --- in each state, we first determine if the subgoal can be considered to be reached and only then do we make the next step. This is a two-stage process, given below. First, we compute the termination condition $\beta(i)$ for each state $i$, according to the following equation.  
\begin{align}
\label{eq-terminate}
\beta_{(k)}(i) \leftarrow \argmax_{\beta_{(k)}(i)\in[0,1]} \; &\beta_{(k)}(i) G(i+1) \;+ \nonumber\\ & \quad (1 - \beta_{(k)}(i)) M_{(k)}(i+1,:) G 
\end{align}
We note that this optimization is of a linear function, therefore we will either have $\beta_{(k)}(i) = 1$ (terminate in state $i$), or $\beta_{(k)}(i) = 0$ (do not terminate in state $i$). Conceptually, this update can be thought of as converting the non-binary subgoal specification $G$ into a binary termination condition $\beta$. Once we have computed $\beta_{(k)}$, we define the diagonal matrix $\beta_{(k)} = \diag(1,\beta_{(k)}(1),\beta_{(k)}(2),\dots, \beta_{(k)}(n)) $ as well as the new matrix $B$ as follows.\footnote{The reader will notice that our matrix $B$ can be understood to be the expected model given the termination condition: $B(\beta_{(k)},M_{k}) = E_{\beta_{(k)}}[I,M_{(k)}]$. However, in our algorithm it is enough to consider it just a matrix.}
\[
B(\beta_{(k)},M_{(k)}) = \beta_{(k)} I + (I - \beta_{(k)}) M_{(k)}
\]
 Here, $I$ is the identity matrix. $B$ summarizes our termination condition --- it behaves like model $M_{(k)}$ for the states where we do not terminate and like the identity model for the states where we do. Once we have this, we can define the actual update, which is executed for each state $i$.
\begin{gather}
\label{eq-act}
a \leftarrow \argmax_a A_a(i+1,:)B(\beta_{(k)},M_{(k)}) G; \nonumber \\ 
M_{(k+1)}(i+1,:) \leftarrow A_a(i+1,:) B(\beta_{(k)},M_{(k)})
\end{gather}
By iterating this many times, we can obtain $M_{\infty}$, which will tend to go from every state to states with high values of the subgoal $G$. The elements of $G$ are specified in the same units as the rewards --- i.e. this algorithm will go, for the non-terminating states, to a state with a particular value of the subgoal if the value of being in the subgoal exceeds the opportunity loss of reward on the way. For the terminating states, the model will still make one step according to the induced policy (see discussion in section \ref{sec-omodels}).

There is one more way we can speed up the algorithm --- through the introduction of \emph{initiation sets}. In this case, instead of selecting an action from the set of all possible actions, we only select an action from the set of allowed actions for a given state (the initiation set). More formally, let $S_a(i)$ be a boolean vector which has `true' in the entries where action $a$ is allowed is state $i$ and `false' otherwise. Equation \ref{eq-act} then becomes the following.
\begin{gather}
\label{eq-act-is}
a \leftarrow \argmax_{a : S_a(i)} A_a(i+1,:) B(\beta_{(k)},M_{(k)}) G; \nonumber \\
M_{(k+1)}(i+1,:) \leftarrow A_a(i+1,:) B(\beta_{(k)},M_{(k)})
\end{gather}
The benefit of using initiation sets is that by not considering irrelevant actions, the whole algorithm becomes much faster. We defer the definition of initiation sets used to section \ref{sec-ep}.

Finally, we solve for several subgoals simultaneously. We use the current state of every model in every iteration, to compute the next iteration for both itself and other models. Denote our subgoals by $G^{(1)},G^{(2)},\dots,G^{(g)}$ and the $k$-th iteration of the models trying to solve these subgoals by $M_{(k)}^{(1)},M_{(k)}^{(2)},\dots,M_{(k)}^{(g)}$. Define the set $\Omega_{(k)}$ as the set of all models (macro-actions) allowed at iteration $k$, i.e. $\Omega_{(k)} = \{A_1,A_2,\dots,A_l,M_{(k)}^{(1)},M_{(k)}^{(2)},\dots,M_{(k)}^{(g)}\}$. This gives rise to the update given below,  for each subgoal $q$ and for each state $i$. We now compute the termination condition.
\begin{align}
\beta_{(k)}^{(q)}(i) \leftarrow \argmax_{\beta_{(k)}(i)\in[0,1]} \; & \beta_{(k)}(i) G^{(q)}(i+1) \; + \nonumber \\ & \quad (1 - \beta_{(k)}(i)) M_{(k)}^{(q)}(i+1,:) G^{(q)} 
\end{align}
The we compute one step of the algorithm according to the equation.
\begin{gather}
\label{eq-act-othermodels}
O \leftarrow \argmax_{O \in \Omega_{(k)}} O(i+1,:) B(\beta_{(k)}^{(q)},M_{(k)}^{(q)}) G^{(q)}; \nonumber \\ \quad M_{(k+1)}^{(q)}(i+1,:) \leftarrow O(i+1,:) B(\beta_{(k)}^{(q)},M_{(k)}^{(q)})
\end{gather}
Solving several subgoals simultaneously can improve the algorithm \cite{options-tl}. The immediate availability of the partial solution to every subgoal leads to faster convergence. In other words, this feature can be used to construct the macro-operator hierarchy at \emph{run time} of the algorithm.\footnote{By this we mean that the option models are built up in run time, possibly using other models. The subgoals are pre-defined and constant.} This is in contrast to many other approaches, where the hierarchy is fixed before the algorithm is run.  

\section{Combining State Aggregation and Options}
\label{sec-aggr}
We saw in section \ref{sec-aggr-mdp} that given the aggregation\footnote{In the work done in this paper, we used hard aggregation so that each row of $\Phi$ contains a one in one place and zeros elsewhere, and the matrix $D$ is a renormalized version of $\Phi^\top$, so that the rows sum to one.} matrix $\Phi$ and the disaggregation matrix $D$, we can convert an action with the transition matrix $P$ and expected reward vector $R$ to an aggregate MDP by using $\tilde{P} = D P \Phi$ and $\tilde{R} = D R$. In our matrix model notation, this becomes as follows.
\begin{align}
\label{compress}
\tilde{A} = \mm{1}{0}{0}{D} A \mm{1}{0}{0}{\Phi}\!\!, \; \text{where} \; A = \mm{1}{0}{R}{P}
\end{align}
This can be viewed as compressing the dynamics, given our aggregation architecture $\Phi$ of size $n \times m$, where $m$ is the number of the aggregate states. We stress that the compressed dynamics define a valid MDP --- therefore the algorithms described in the previous section are \emph{convergent}. 

The main idea of our algorithm is the following: define a subgoal, solve it (i.e. obtain a model for reaching it) and then add it to the action set of the original problem and use it as a macro-action, gaining speed. We repeat this for many subgoals. Solving subgoals is fast because we do it in the small, aggregate state space. To be precise, we pick a subgoal $\tilde{G}$ (see section \ref{experiments} for examples) and an approximation architecture $\Phi$. We then compress our actions with eq.\ \ref{compress} and use compressed actions in VI according to eqs. \ref{eq-terminate} and \ref{eq-act}. This gives us a model $\tilde{M}_\infty$ solving the subgoal in the aggregate state space. We want to use this model to help solve the original MDP.  

However, we cannot do this directly since our model $\tilde{M}_\infty$ is defined with respect to the \emph{aggregate} state space and has size $(m+1) \times (m+1)$ --- we need to find a way to convert it to a model defined over the original state space, of size $(n+1) \times (n+1)$. The new model also has to be \emph{valid}, i.e.\ correspond to a sequence of actions.\footnote{That is why it is not possible to just upscale the model by writing:
$
\mm{1}{0}{0}{\Phi} \tilde{M}_\infty \mm{1}{0}{0}{D}
$.}
 
The idea is to make the following transformation: from the aggregate model, we compute the option in the aggregate state space, we then up-scale the option to the original state space, construct a one-step model and then construct the long-term model from it. Concretely, we first compute the \emph{option} corresponding to the model $\tilde{M}_\infty$. The option consists of the policy $\mu$ and the termination condition $\beta$. We obtain the termination condition by using eq.\ \ref{eq-terminate} for the aggregate states. The policy $\mu$ is obtained greedily for each aggregate state $x$.
\begin{align}
\mu(x) = \argmax_{c} \tilde{A_c}(x+1,:) B(\beta,\tilde{M}_{\infty}) \tilde{G}
\end{align}
Now, we can finally build a one-step model in terms of the original state-space. We do this according to the following equation, which we use for each state $i$.
\begin{align}
\label{upscale1}
 M'(i+1,:) \;= \; &\left(1-\beta(\phi(i))\right) \;A_{\mu(\phi(i))}(i+1,:) \; + \nonumber \\ & \quad \beta(\phi(i)) \; I(i+1,:) 
\end{align}
In the above, we denote by $I$ the identity matrix of size $(n+1) \times (n+1)$ and by $\phi(i)$ the aggregate state corresponding\footnote{Note that the equation could be easily generalized to the case where the aggregation is soft --- i.e. there are several aggregate states corresponding to $i$, simply by summing all the possibilities as weighted by the aggregation probabilities.} to the original state $i$. In more understandable terms, $M'$ has rows selected by the policy $\mu$ wherever the option does not terminate and rows from the identity matrix wherever it does. Now, we do not just need a model that takes us one step towards the subgoal --- we want one that takes us all the way. Therefore, we continually evaluate the option by exponentiating the model matrix, producing $M'^\infty$. Now, this new model still has rows from the identity matrix where the option terminates --- therefore it does not correspond to a valid combination of primitive actions. To solve this problem, we compute $M''$, according to the following equation (for each state $i$).
\begin{align}
\label{upscale3}
M''(i+1,:) \; = \; (1-\beta(\phi(s))) \; M'^
\infty(i+1,:) \; + \nonumber \\ \quad \beta(\phi(s)) \; A_{\mu(\phi(s))}(i+1,:)  
\end{align} 
$M''$ contains rows from $M'^\infty$ where the option does not terminate and rows dictated by the option policy where it does. This guarantees it is a valid combination of primitive actions and can be added to the action set and treated like any other action. We now run value iteration (equation \ref{eq-vi}) using the extended action set --- the original actions and the subgoal models $(M'')^{(q)}$ corresponding to each subgoal $q$. This is s faster than using the original actions alone, even after factoring in the time used to compute the subgoal models (see section \ref{experiments}).

\begin{obs}
Value Iteration with the action set $\mathcal{A} \cup \{ (M'')^{(1)}, \dots, (M'')^{(g)}\}$ converges to the optimal value function of the MDP.
\end{obs}

\begin{proofoutline}
The addition of subgoal macro-operators to the action set does not change the fixpoint of value iteration because the macro-operators are, by construction, compositions of the original actions. See supplement to existing work \cite{options-tl} for a formal proof of a more general proposition. 
\end{proofoutline}

This observation tells us that our algorithm will always exactly solve the MDP, computing $V^\star$. The worst thing that can happen is that the subgoal macro-operators will be useless i.e. the resulting value iteration will take as many iterations as without them. 

\section{Why not Use Linear Features}
\label{sec-no-lin}
Looking at eqn. \ref{compress} one may ask if this is the best way to compress actions. It may seem that using \emph{linear} features \cite{approx-valiteration,lizotte-fvi,vroy-avi} may be better because they are more expressive and easier to come up with than $\Phi$ and $D$. Specifically, consider the following way of compressing actions, as an alternative to eq.\ \ref{compress}. Define the approximation architecture $\breve{V} = \qPd w$ for modelling value functions, the sequence of which will converge to the optimal value function. We begin by defining the projection operator \cite{Parr2008,SorgSinghLinOptions} that compresses a table-lookup model $M$ into a model that works with linear features,
\begin{align}
\label{eamodel}
\qam(M) = \modelmatrix{0}{\qProjm} M \modelmatrix{0}{\qPd}
\end{align}
In the above, $\qProjm = (\qPd^\top \qXa \qPd)^{-1} \qPd^\top \qXa$, and $\qXa$ is a diagonal matrix with entries corresponding to a distribution over the original states of the MDP. We introduce names for the minor matrices of the models: $\qam(M) = \modelmatrix{q}{F}$ and $M = \modelmatrix{R}{P}$. We note that eq.\ \ref{eamodel} ensures that the model $\qam(M)$ is the best approximation of the model $M$ in the sense that it solves the  optimization problems:\footnote{The norms are defined in the following way: $\| V \|_\qXa = \sqrt{V^\top \qXa V}$ and $ \| A \|_\qXa =  \sqrt{\operatornamewithlimits{trace} \left( A^\top \qXa A \right). }$} $F = \argmin_F \|\qPd F - P\qPd \|_\qXa$ and $q = \argmin_q \| \qPd q - R\|_\qXa $. In the above, the optimization is applied to the transition and reward components separately; also, each column of $F$ is treated independently of the others. The semantics of the above is as such: each column $k$ of $F$ should be such as to make the entry $s$ of the corresponding $k$-th column of $\qPd F$ as close as possible to the feature number $k$ of the next state, where the index of the current state is $s$. Similarly, $\qPd q$ is picked so as to approximate the expected next reward for each state. In other words, $F$ is a linear dynamical system that models the one-step dynamics on features of the Markov chain corresponding to an action. One might hope that this $F$ and $q$ linear dynamical system could be used in much the same way as the MDP compressed with state aggregation to $\tilde{P}$ and $\tilde{R}$.

But there is a problem with the compressed models defined according to eq.\ \ref{eamodel}. Consider an action with the transition matrix and approximation architecture $\qPd$ given below.
{\[
P = \gamma \hbox{\tiny $ \left [ \begin{array}{cccc} 1 & 0 & 0 & 0  \\ 1 & 0 & 0 & 0 \\0 & 1 & 0 & 0 \\ 0 & 0 & 1 & 0 \\ \end{array} \right ] $}; \quad
\qPd = \hbox{\tiny $ \left [ \begin{array}{cc} 1 & 1  \\ 1 & 0 \\ 0 & 1 \\ 0 & 0 \\ \end{array} \right ] $}; \quad F = \gamma \frac13 \left [ \begin{array}{cc} 2 & 3  \\ 2 & 0 \end{array} \right ]
\]}
It can be easily shown that this, paired with a uniform distribution $\qXa$, produces the matrix $F$ given above. But this matrix has spectrum outside of the unit circle for some $\gamma < 1$ --- hence if this action is composed with itself time and again, the VI algorithm will diverge. The argument given above shows that we cannot\footnote{If $\qXa$ is chosen to be the diagonal matrix with entries from the left principal eigenvector of the state transition matrix of the $P$, it turns out \cite{approx-valiteration} that, the matrix $F$ has spectrum within the unit circle, which leads to a convergent algorithm. However, the problem is that this approach only takes into account the long-term effects of actions. For instance in a two-dimensional grid-world, for the action that goes right, such a distribution will be non-zero only along the rightmost edge of the grid-world. In practice $P$ will seldom be ergodic and all information in the non-recurrent part will be lost. This cannot lead to good overall solutions of the MDP.} use eq.\ \ref{eamodel} for \emph{arbitrary} features $\qPd$ and distributions $\qXa$. On the other hand, our framework based on eq.\ \ref{compress} does not suffer from the described divergent behaviour. Also, it does not depend on any distribution over the states, meaning there is one less parameter to the algorithm. 

\section{Experiments}
\label{experiments}
\begin{figure}[t]
\caption{Run-times of our algorithm, plain VI and model VI. All algorithms compute $V^\star$.}
\begin{center}
{ \small 
\begin{tabular}{ l c c c }
  & & & \bf options + \\
  \bf Domain & \bf plain VI & \bf model VI & \bf aggr.  \\
  Taxi (determ.) & 6.43 s. & 11.64 s. & 4.57 s. \\
  Taxi (stoch.) & 8.30 s. & 47.80 s. & 4.83 s. \\ 
  Hanoi (determ.) & 23.45 s. & 51.65 s. & 11.57 s. \\
  Hanoi (stoch.) & 27.31 s. & 357.52 s. & 21.71 s. \\
  8-puzzle (determ.) & 100.19 s. & 221.20 s. & 85.94 s.  
\end{tabular}
}
\end{center}
\label{res-fig}
\vspace*{-12pt}
\end{figure}
We applied our approach to three domains: Taxi, Hanoi and 8-puzzle. In each case we compared several variants of VI, including our approach combining state aggregation and options. For vanilla VI we considered algorithms based on both eq.\ \ref{eq-vi} (the familiar algorithm, denoted plain VI) and eq.\ \ref{eq-vi-model} (model VI, where complete models are constructed). Figure \ref{res-fig} summarises the  solution times for each domain; more details are given in the following domain-specific subsections. We, however, stress beforehand that our algorithm produced a speeed-up for each of the domains we tried.

\subsection{The TAXI Problem}
\label{exp-taxi}
TAXI \cite{dietterich1998maxq} is a prototypical example of a problem which combines spatial navigation with additional variables. Denote the number of states as $n$ (here $n = 7000 + 1$) and the number of aggregate states as $m$ (here $m = 25 + 1$). The one state is the sink state.

In our first experiment, we ran four algorithms computing the same optimal value function., one for each combination of using (or not) state aggregation and options. Consider using neither aggregation nor options --- this is model VI, one iteration of which has a complexity of O($n^2|\mathcal{A}| + n^3$), in practice it is O($n|\mathcal{A}|$) because of sparsity. It takes 22 iterations to complete. Now consider the version with subgoals but no aggregation. Here, we have 5 subgoals: one for getting to each pick-up location or the fuel pump. An iteration now has complexity O($g ((|\mathcal{A}| + g) n^2 + n^3) $). Because of sparsity, this becomes O($g ((|\mathcal{A}| + g) n + n) $) = O($g ((|\mathcal{A}| + g) n$). The algorithm needs 8 iterations less to converge, because subgoals allow it to make jumps. However, due to the increased cost of each iteration, the time required to converge increased. Now look at the version with aggregation (see section \ref{sec-aggr}) and no options. There are 26 aggregate states. We map each original state to one of 25 states by taking the taxi position and ignoring other variables. Sink state (state 7001) gets mapped to the aggregate sink state (state 26). We proceed in two stages. First, all actions are compressed (eq.\ \ref{compress}). Then, the problem is solved using model VI in this smaller state-space. This takes 330 iterations, but is fast because $m$ is small --- the complexity is O($m^2|\mathcal{\tilde{A}}| + m^3$). We then obtain the value function of the aggregate system and upscale it, then we use the new value function to obtain a greedy model (i.e. each row comes from the action that maximizes that row times $\bar{V}$), which we use as initialization in our iteration, which takes 3 iterations less than our original algorithm. Now consider the final version, where the benefits of aggregation and options are combined. Again, the algorithm consists of two stages. First, we use compressed actions to compute models for getting to the five subgoals. This requires 17 iterations; the complexity of each is O($g ((|\mathcal{\tilde{A}}| + g) m^2 + m^3) $), where $g=5$. This is fast since $m$ is small. We now upscale these models. We see that if we add the five macro-actions, we do not need the original four actions for moving, as all sensible movement is to one of the five locations. The algorithm now takes only 7 iterations to converge.\footnote{We need an iteration to: (1) go to the fuel pump, (2) fill in fuel, (3) go to passenger, (4) pick up passenger, (5) go to destination, (6) drop off passenger. The 7th iteration comes from the termination condition.} The run-time\footnote{This is slightly different from the result in fig. \ref{res-fig} since after the models have been upscaled, we can proceed either with plain VI (as is fig. \ref{res-fig}) or with model VI, which we do here to make the comparison fair.} is 6.55 s, i.e.\ a speed-up of 1.8 times over model VI. Results for all four versions are summarized in figure \ref{res-taxi}. We also constructed a stochastic version of the problem, with a probability of 0.05 of staying in the original state when moving. Results are qualitatively similar and are in figure \ref{res-taxi}. The speed-up from combining options with aggregation was greater at 7.1 times. We stress the main result.\footnote{If the number of subgoals and actions is constant.} In the deterministic case, we replace many O($n$) iterations with many O($m^3$) iterations followed by few O($n$) iterations. For stochastic problems, we replace many O($n^3$) iterations with many O($m^3$) iterations followed by few O($n^3$) iterations.

\begin{figure}[t]
\caption{Run-times of the algorithm in the deterministic and stochastic versions of TAXI .}
\vspace{10pt}

       \begin{center}
		{\small
		\begin{tabular}{r|p{2cm}|p{2cm}}
		deter. & \bf no aggreg. & \bf aggregation \\ \hline
		\bf no options & 22 iter. & 330 + 19 iter. \\
		& 11.64 s. & 11.73 s. \\ \hline
		\bf options & 14 iter. & 17 + 7 iter. \\
		& 78.20 s. & 6.55 s. \\
		\end{tabular}
		}
	   \end{center}

		\begin{center}
		{\small
		\begin{tabular}{r|p{2cm}|p{2cm}}
		stoch. & \bf no aggreg. & \bf aggregation \\ \hline
		\bf no options & 30 iter. & 331 + 28 iter. \\
		& 47.80 s. & 26.04 s. \\ \hline
		\bf options & 18 iter. & 20 + 7 iter. \\
		& 256.04 s. & 6.78 s. \\
		\end{tabular}
		}
		\end{center}    
\label{res-taxi}
\vspace*{-12pt}
\end{figure}

In our second experiment, as a digression from the main thrust of the paper, we tried a different approach: we can use the aggregation framework to compute an \emph{approximate} value function, gaining speed. Our actions are compressed as defined by eq. \ref{compress}, and we simply apply eq.\ \ref{eq-vi}. This process gives us a value function $\tilde{V}^\star$ defined over the aggregate state space (in the first case we need to extract it from the reward part of the model). We upscale this value function to the original states using the equation $
\bar{V} = \mm{1}{0}{0}{\Phi} \tilde{V}^\star $. Of course, the obtained value function $\bar{V}$ is only \emph{approximately} optimal in the original problem. Consider a $\Phi$ with 501 aggregate states --- the aggregation happens by eliminating the fuel variable and leaving others intact. The algorithm used is given by eq.\ \ref{eq-vi-model}, applied to compressed actions. It takes 2.94 s / 28 iterations to converge (determ.) and 3.08 s / 30 iterations (stoch.). The learned value function corresponds to a policy which ignores fuel, never visits the pump, but otherwise, if there is enough fuel, transports the passenger as intended. We have shown an important principle --- if we have an aspect of a system that we feel our solution can ignore, we can eliminate it and still get an \emph{approximate} solution. The benefit is in the speed-up. --- in our case, with respect to solving the original MDP using plain VI, it is 2.2 (determ.) / 2.7 (stoch.).

\subsection{The Towers of Hanoi}
 For $r$ disks, our state representation in the Towers of Hanoi is an $r$-tuple, where each element corresponds to a disk and takes values from $\{1,2,3\}$, denoting the peg.\footnote{Note that the state representation itself disallows placing a larger disk on top of a smaller one.} There are three actions, two for moving the smallest disk and one for moving a disk between the remaining two pegs. It is known that VI for this problem takes $2^r$ iterations to converge. To speed up the iteration, we introduced the following state abstraction. There are $r-2$ sub-problems of size $2$,...,$r-1$. First, we solve the problem with $2$ disks, i.e. our abstraction only considers the position of the two smallest disks, ignoring the rest. There are three subgoals, one for placing the two disks on each of the pegs. Then, once we obtained three models for the subgoals, we use them to solve the sub-problem of size $3$, ignoring all disks except the three smallest ones. Again, there are three subgoals. We proceed until we solve the problem with $r$ disks. For each subgoal, we  need 4 iterations (Three moves and the 4th is required for the convergence criterion). The total number of iterations is $4\times 3 \times r$, i.e. it is linear in the state space. For 8 disks this means the following speed-up: 11.57 s (with subgoals) vs. 51.65 s (model VI) vs. 23.45 s (plain VI). We note however, that the time complexity of the algorithm with subgoals is still exponential in $r$, because whereas the number of iterations is only linear, in each iteration we need to iterate the whole state space, which is exponential.\footnote{However, this problem is not particular to our approach --- every algorithm that purports to compute the value function \emph{for each state} will have computational complexity at least as high as the number of such states.} For a stochastic version, the run-times were 357.52 s for model VI, 27.31 s for plain VI and 21.71 s for computing the same optimal value function with options with aggregation.

\subsection{The 8-puzzle}
\begin{figure}[t]
\caption{The subgoal used and run-times for the 8-puzzle. All algorithms compute $V^\star$.}
\vspace{10pt}
    \begin{center}
	\begin{minipage}{0.4\linewidth}
		\begin{center}
		\small
		\begin{tabular}{r c c c}
		 & \bf iter. & \bf time elapsed \\ 
		model VI & 32 & 221.20 s. \\
		plain VI & 33 & 100.19 s. \\
		subgoal & 25 & 109.51 s. \\
		subgoal w. init. set & 25 & 85.94 s.
		\end{tabular}
		\end{center}  
  	\end{minipage} 
	\begin{minipage}{0.15\linewidth}
       \begin{center}
		{
		\vspace*{-10pt}
		\def\svgwidth{1.3cm}
		\small 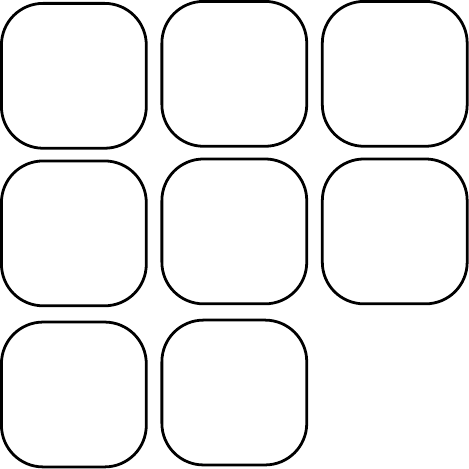
		\vspace*{-12pt}
		}
		\end{center}
  	\end{minipage}
  	\end{center} 
\label{res-ep}
\end{figure}

\label{sec-ep}
The 8-puzzle \cite{story1879notes, reinefeld1993complete} is well-known in the planning community. Our subgoal is shown in figure \ref{res-ep}.\footnote{Other subgoals are shown in the documentation accompanying the source code. Please also consult the source code, where all subgoals are implemented.} `A',`B', and `C' denote groups of tiles. The subgoal consists in arranging the tiles so that each group is in correct place (but tiles \emph{within} each group are allowed to occupy an incorrect place). The matrix $\Phi$ is such that the original configuration of the tiles is mapped onto one where each tile is only marked with the group it belongs to. Using the subgoal alone did not result in a speed-up, so we used the notion of initiation sets \cite{Sutton99betweenmdps}. We trained the subgoal for 9 iterations (the number 9 was obtained by trial and error), so the obtained model is only able to reach the subgoal for some starting states (the ones at most 9 steps away from the subgoal in terms of primitive actions). We upscaled the model, but this time the new model had an initiation set containing only those states from which the subgoal is reachable. The iteration we then used is plain value iteration, extended to initiation sets. The intuition behind initiation sets is that it only makes sense to use a subgoal if we are already in a part of the state space close to it. Thus, we obtained a total run-time of 85.94 seconds, which amounts to a speed-up of 1.17 over plain value iteration. The results are in figure \ref{res-ep}. 

\section{Conclusions}
We introduced new Bellman optimality equations that facilitate VI with options. These equations can be combined with state aggregation in a sound way, and therefore can be applied to the solution of medium-sized MDPs.\footnote{We provide software used in our experiments under GPL in the hope that others may use it for their problems.} This is the first algorithm combining options and state abstraction which is \emph{guaranteed to converge}. This is notable because other proposed approaches, notably based on linear features, are known to diverge even for small problems. Finally, we have shown experimentally that the benefits of options and state aggregation are only realized when they are applied \emph{together}.

\bibliographystyle{abbrv}
\bibliography{ucl-bb}  

\section{Appendix}
In this appendix, we discuss background information concerning state aggregation for MDPs, adapted to the notation of our paper. This is necessary because Bertsekas' original notation is difficult to apply to our work. We stress that the ideas presented in this appendix are entirely due to Bertsekas \cite{bertsekas-vol2}.  

We are concerned with an MDP which has $|\mathcal{A}|$ actions, and for an action $a$ the probability transition matrix is $P_a$, defined by $P_a(i,j) = \gamma \text{Pr}(i_{t+1} = j | i_{t} = i, a_t = a)$ and the vector of expected rewards for each state is $R_a$, defined by $R_a(i) = \text{E} \left[r_{t}  |  i_{t} = i, a_t = a \right]$. There are $m$ aggregate states. In addition, we introduce two matrices\footnote{We employ the names introduced by Bertsekas \cite{bertsekas-vol2}.} defining the approximation architecture: the \emph{aggregation matrix} $\Phi$ and the \emph{disaggregation matrix} $D$. The matrix $\Phi$ has dimensions $n \times m$ and the matrix $D$ has dimension $m \times n$. It is useful to think about these matrices as conversion operators: the matrix $\Phi$ converts a value function defined over the aggregate states into one defined over the original states; conversely, the matrix $D$ converts a value function defined over the original states into one defined over the aggregate states. There are no conditions on these matrices other than the rows have to sum to one, as they are probability distributions modeling, for $\Phi$, the degree by which each state is represented by various aggregate states and, for $D$, the degree to which a certain aggregate state corresponds to various original states. Having defined the matrices, we can define our first approximation step. The Bellman optimality operator in the original MDP is called $T$, and is defined by $(TV)(i) = \max_a (P_a V)(i) + R_a(i)$ and the optimum value function $V^\star$ satisfies the fixpoint equation $V^\star = T V^\star$. Now, the approximation consists in solving the following equation instead (we will see later that this is not solved exactly and further approximation is necessary).
\begin{align}
\label{eq-aggr-bellman}
\Vb^\star = DT(\Phi \Vb^\star)
\end{align}
In the above, we use $ \tilde{\cdot} $ to denote the aggregate problem. We note that this equation operates on a shorter value function --- $\Vb^\star$ has entries corresponding to \emph{aggregate} states. The idea is, of course that the number of aggregate states is tractable, so we can compute $\Vb^\star$. However, we need to reformulate the equation since in its present form it contains the operator $T$, which still operates on the original states. To do so, we expand the definition of $T$, to obtain the following state-wise equation, for the aggregate state $x$.
\[ 
\Vb^\star(x) = \sum_i d_{xi} \left( \max_{a} P_a(i,:) \Phi \Vb^\star + R_a(i) \right)
\]   
This equation leads to the following iterative algorithm, which computes $\Vb^\star$ as $k \rightarrow \infty$.
\[ 
\Vb_{(k+1)}(x) = \sum_i d_{xi} \left( \max_{a} P_a(i,:) \Phi \Vb_{(k)} + R_a(i) \right)
\]   
In the above, $P_a(i,:)$ denotes the row number $i$ of the probability transition matrix corresponding to action $a$ (in terms of the original states). Value functions are assumed to be column vectors. In order to be able to operate exclusively with objects that have dimensionality corresponding to the number of \emph{aggregate} states, we introduce another approximation and namely we do the following.
\[ 
\Vb_{(k+1)}(x) = \max_{a} \sum_i d_{xi} \left( P_a(i,:) \Phi \Vb_{(k)} + R_a(i) \right)
\]   
We note that this approximation is exact if states mapping to a single aggregate state all have the same optimal action. Now, we can reformulate the equation in the following way.
\begin{align}
\label{eq-maxout} 
\Vb_{(k+1)}(x) &= \max_{a} D(x,:) P_a \Phi \Vb_{(k)} + D(x,:) R_a \nonumber \\&= \max_{a} (\tilde{P}_a \Vb_{(k)})(x) + \tilde{R}_a(x)
\end{align}   
In the above, $D(x,:)$ denotes the row of $D$ corresponding to aggregate state $x$ and $P_a$ is the probability transition matrix corresponding to action $a$ in the original MDP. Now, we note that solving the above equation is equivalent to solving a modified MDP with actions corresponding to the original actions, probability transition matrices given by $\tilde{P}_a = D P_a \Phi$ and expected reward vectors given by $\tilde{R}_a = D R_a$. The states of the modified MDP are the aggregate states.

Therefore, under our two explained approximations, solving the original MDP may be replaced by solving a much smaller aggregate MDP, by computing $\tilde{P}_a$ and $\tilde{R}_a$. The solution can then be computed by any known algorithm, although in this paper we focus only on VI. We emphasize that the VI is \emph{convergent} because the matrices $\tilde{P}_a$ and $\tilde{R}_a$ define a valid MDP. We stress again that this involves two approximations: first, we are solving a modified Bellman equation \ref{eq-aggr-bellman} that utilizes state aggregation and second, we move the $\max$ operator outside of the sum in equation \ref{eq-maxout}.

\end{document}

%% file: subgoal10.pdf_tex
\begingroup%
  \makeatletter%
  \providecommand\color[2][]{%
    \errmessage{(Inkscape) Color is used for the text in Inkscape, but the package 'color.sty' is not loaded}%
    \renewcommand\color[2][]{}%
  }%
  \providecommand\transparent[1]{%
    \errmessage{(Inkscape) Transparency is used (non-zero) for the text in Inkscape, but the package 'transparent.sty' is not loaded}%
    \renewcommand\transparent[1]{}%
  }%
  \providecommand\rotatebox[2]{#2}%
  \ifx\svgwidth\undefined%
    \setlength{\unitlength}{134.9406412bp}%
    \ifx\svgscale\undefined%
      \relax%
    \else%
      \setlength{\unitlength}{\unitlength * \real{\svgscale}}%
    \fi%
  \else%
    \setlength{\unitlength}{\svgwidth}%
  \fi%
  \global\let\svgwidth\undefined%
  \global\let\svgscale\undefined%
  \makeatother%
  \begin{picture}(1,0.99946462)%
    \put(0,0){\includegraphics[width=\unitlength]{subgoal10.pdf}}%
    \put(0.08012737,0.75122597){\color[rgb]{0,0,0}\makebox(0,0)[lb]{\smash{A}}}%
    \put(0.42255811,0.75541173){\color[rgb]{0,0,0}\makebox(0,0)[lb]{\smash{A}}}%
    \put(0.76498881,0.75541173){\color[rgb]{0,0,0}\makebox(0,0)[lb]{\smash{A}}}%
    \put(0.08012737,0.41494055){\color[rgb]{0,0,0}\makebox(0,0)[lb]{\smash{B}}}%
    \put(0.42255811,0.4191263){\color[rgb]{0,0,0}\makebox(0,0)[lb]{\smash{B}}}%
    \put(0.76498881,0.4191263){\color[rgb]{0,0,0}\makebox(0,0)[lb]{\smash{B}}}%
    \put(0.08012737,0.07108567){\color[rgb]{0,0,0}\makebox(0,0)[lb]{\smash{C}}}%
    \put(0.42255812,0.07527139){\color[rgb]{0,0,0}\makebox(0,0)[lb]{\smash{C}}}%
  \end{picture}%
\endgroup%

%% file: OptionsAggregationPaper.bbl
\begin{thebibliography}{10}

\bibitem{andre2002state}
D.~Andre and S.~J. Russell.
\newblock State abstraction for programmable reinforcement learning agents.
\newblock In {\em AAAI Conference on Artificial Intelligence / Annual
  Conference on Innovative Applications of Artificial Intelligence}, pages
  119--125, 2002.

\bibitem{Bellman:1957}
R.~Bellman.
\newblock {\em Dynamic Programming}.
\newblock Princeton University Press, Princeton, NJ, USA, 1957.

\bibitem{bertsekas-vol2}
D.~P. Bertsekas.
\newblock {\em Dynamic Programming and Optimal Control}, volume~2.
\newblock Athena Scientific Belmont, 2012.

\bibitem{approx-valiteration}
D.~P. De~Farias and B.~Van~Roy.
\newblock On the existence of fixed points for approximate value iteration and
  temporal-difference learning.
\newblock {\em Journal of Optimization Theory and Applications}, 105:589--608,
  2000.

\bibitem{dietterich1998maxq}
T.~G. Dietterich.
\newblock {The MAXQ Method for Hierarchical Reinforcement Learning.}
\newblock In {\em International Conference on Machine Learning}, pages
  118--126, 1998.

\bibitem{hengst2002discovering}
B.~Hengst.
\newblock { Discovering hierarchy in reinforcement learning with HEXQ }.
\newblock In {\em International Conference on Machine Learning}, volume~2,
  pages 243--250, 2002.

\bibitem{jong2005state}
N.~K. Jong and P.~Stone.
\newblock { State Abstraction Discovery from Irrelevant State Variables. }.
\newblock In {\em International Joint Conferences on Artificial Intelligence},
  pages 752--757, 2005.

\bibitem{jonsson2001automated}
A.~Jonsson and A.~G. Barto.
\newblock {Automated state abstraction for options using the U-tree algorithm}.
\newblock {\em Advances in neural information processing systems}, pages
  1054--1060, 2001.

\bibitem{konidaris2007building}
G.~Konidaris and A.~G. Barto.
\newblock { Building Portable Options: Skill Transfer in Reinforcement
  Learning. }.
\newblock In {\em International Joint Conferences on Artificial Intelligence},
  volume~7, pages 895--900, 2007.

\bibitem{korf1985learning}
R.~Korf.
\newblock {\em Learning to Solve Problems by Searching for Macro-Operators}.
\newblock Research Notes in Artificial Intelligence, Vol 5. Pitman, 1985.

\bibitem{lizotte-fvi}
D.~J. Lizotte.
\newblock Convergent fitted value iteration with linear function approximation.
\newblock In J.~Shawe-Taylor, R.~Zemel, P.~Bartlett, F.~Pereira, and
  K.~Weinberger, editors, {\em Advances in Neural Information Processing
  Systems 24}, pages 2537--2545. 2011.

\bibitem{Parr2008}
R.~Parr, L.~Li, G.~Taylor, C.~Painter-Wakefield, and M.~L. Littman.
\newblock An analysis of linear models, linear value-function approximation,
  and feature selection for reinforcement learning.
\newblock In {\em Proceedings of the 25th international conference on Machine
  learning}, ICML '08, pages 752--759, New York, NY, USA, 2008. ACM.

\bibitem{raey}
D.~Precup, R.~S. Sutton, and S.~Singh.
\newblock Theoretical results on reinforcement learning with temporally
  abstract options.
\newblock In {\em Machine Learning: ECML-98}, volume 1398 of {\em Lecture Notes
  in Computer Science}, pages 382--393. Springer Berlin Heidelberg, 1998.

\bibitem{reinefeld1993complete}
A.~Reinefeld.
\newblock Complete solution of the eight-puzzle and the benefit of node
  ordering in ida*.
\newblock In {\em International Joint Conference on Artificial Intelligence},
  pages 248--253, 1993.

\bibitem{ribas2011neural}
J.~J. Ribas-Fernandes, A.~Solway, C.~Diuk, J.~T. McGuire, A.~G. Barto, Y.~Niv,
  and M.~M. Botvinick.
\newblock A neural signature of hierarchical reinforcement learning.
\newblock {\em Neuron}, 71(2):370--379, 2011.

\bibitem{options-tl}
D.~Silver and K.~Ciosek.
\newblock Compositional planning using optimal option models.
\newblock In {\em 29th International Conference on Machine Learning}, 2012.

\bibitem{SorgSinghLinOptions}
J.~Sorg and S.~Singh.
\newblock Linear options.
\newblock In {\em Proceedings of the 9th International Conference on Autonomous
  Agents and Multiagent Systems: Volume 1 - Volume 1}, AAMAS '10, pages 31--38,
  Richland, SC, 2010. International Foundation for Autonomous Agents and
  Multiagent Systems.

\bibitem{story1879notes}
W.~E. Story.
\newblock Notes on the ``15'' puzzle.
\newblock {\em American Journal of Mathematics}, 2(4):397--404, 1879.

\bibitem{Sutton99betweenmdps}
R.~Sutton, D.~Precup, and S.~Singh.
\newblock {Between MDPs and Semi-MDPs: A Framework for Temporal Abstraction in
  Reinforcement Learning}.
\newblock {\em Artificial Intelligence}, 112:181--211, 1999.

\bibitem{Sutton95tdmodels:}
R.~S. Sutton.
\newblock { TD Models: Modeling the World at a Mixture of Time Scales }.
\newblock In {\em Proceedings of the Twelveth International Conference on
  Machine Learning}, pages 531--539. Morgan Kaufmann, 1995.

\bibitem{vroy-avi}
B.~{Van Roy}.
\newblock { TD(0) Leads to Better Policies than Approximate Value Iteration }.
\newblock In Y.~Weiss, B.~Sch\"{o}lkopf, and J.~Platt, editors, {\em Advances
  in Neural Information Processing Systems 18}, pages 1377--1384. 2005.

\bibitem{wang-maxq}
H.~Wang, W.~Li, and X.~Zhou.
\newblock Automatic discovery and transfer of maxq hierarchies in a complex
  system.
\newblock In {\em ICTAI}, pages 1157--1162, 2012.

\end{thebibliography}
